\documentclass[letterpaper, 10 pt, conference]{ieeeconf}  

\IEEEoverridecommandlockouts                              

\usepackage{amsmath,amsfonts}
\usepackage{algorithmic}
\usepackage{algorithm}
\usepackage{array}
\usepackage[caption=false,font=normalsize,labelfont=sf,textfont=sf]{subfig}
\usepackage{textcomp}
\usepackage{stfloats}
\usepackage{url}
\usepackage{verbatim}
\usepackage{graphicx}
\usepackage{cite}
\usepackage[breaklinks,hidelinks]{hyperref}
\usepackage{cleveref}
\usepackage{siunitx}
\usepackage[nolist,nohyperlinks]{acronym}
\usepackage{booktabs}
\usepackage{makecell}
\usepackage{multirow}
\acrodef{rl}[RL]{Reinforcement Learning}
\acrodef{mlp}[MLP]{Multi-Layer Perceptron}
\acrodef{lstm}[LSTM]{Long Short-Term Memory}
\acrodef{fov}[FOV]{Field-of-View}
\acrodef{appo}[APPO]{Asynchronous Proximal Policy Optimization}
\acrodef{rnn}[RNN]{Recurrent Neural Network}
\acrodef{gru}[GRU]{Gated Recurrent Unit}
\acrodef{pomdp}[POMDP]{Partially Observable Markov Decision Process}
\acrodef{nbv}[NBV]{Next-Best-View}
\acrodef{imu}[IMU]{Inertial Measurement Unit}
\acrodef{vae}[VAE]{Variational Autoencoder}
\acrodef{dce}[DCE]{Deep Collision Encoder}

\usepackage[table]{xcolor}

\title{\LARGE \bf
Reinforcement Learning for Active Perception \\
in Autonomous Navigation
}

\author{Grzegorz Malczyk$^{*}$, Mihir Kulkarni and Kostas Alexis%
\thanks{This work was supported by the Research Council of Norway under Award NO-338694 and the Horizon Europe Grant Agreement No. 101119774. The authors are with the Department of Engineering Cybernetics, Norwegian University of Science and Technology (NTNU), Norway.}
\thanks{$^*$Corresponding author. Email: \tt\small grzegorz.malczyk@ntnu.no}
}

\begin{document}

\maketitle
\thispagestyle{empty}
\pagestyle{empty}

\begin{abstract}
This paper addresses the challenge of active perception within autonomous navigation in complex, unknown environments. Revisiting the foundational principles of active perception, we introduce an end-to-end reinforcement learning framework in which a robot must not only reach a goal while avoiding obstacles, but also actively control its onboard camera to enhance situational awareness. The policy receives observations comprising the robot state, the current depth frame, and a particularly local geometry representation built from a short history of depth readings. To couple collision-free motion planning with information-driven active camera control, we augment the navigation reward with a voxel-based information metric. This enables an aerial robot to learn a robust policy that balances goal-directed motion with exploratory sensing. Extensive evaluation demonstrates that our strategy achieves safer flight compared to using fixed, non-actuated camera baselines while also inducing intrinsic exploratory behaviors. 
\end{abstract}
\vspace{-1ex}
\section{INTRODUCTION}
Autonomous aerial robots are increasingly deployed in complex and cluttered environments, where safe navigation and effective perception are critical for missions such as infrastructure inspection, search and rescue, and environmental monitoring. Traditionally, robot missions involve a sequence of waypoints with robots tasked to reach these targets while avoiding obstacles. During such point-to-point navigation, perception is often treated as a passive process ``simply'' consuming data collected during the motion of the robot while the sensors are fixed on it. Actuated cameras remain rare in navigation research. This separation, however, overlooks a crucial fact: perception itself is an active process. It involves decisions about what, when, and where to sense to improve situational awareness and task performance~\cite{chen2011active}.

The concept of active perception, early articulated in \cite{aloimonos1987active,bajcsy1988active} and recently revisited in \cite{bajcsy2018revisiting}, emphasizes that sensing should be purpose-driven. Early robotic systems~\cite{chen2011active} demonstrated the potential of foveated cameras and movable sensors to improve task-relevant perception, but these approaches were limited by the hardware and computational constraints of the time.
Building on these ideas, more recent work in \ac{nbv} planning and active mapping \cite{vasquez2014volumetric,bircher2016receding} has focused on selecting viewpoints that maximize information gain or coverage. However, these approaches primarily target mapping and exploration objectives with limited attention to optimizing viewpoint selection for navigation toward specific goals. Even in methods that formulate intrinsic attention objectives, such as the work in ~\cite{dang2018visual}, most literature is limited to non-actuated cameras that are fixed on the robot's frame.
\begin{figure}
    \centering
    \includegraphics[width=0.98\columnwidth]{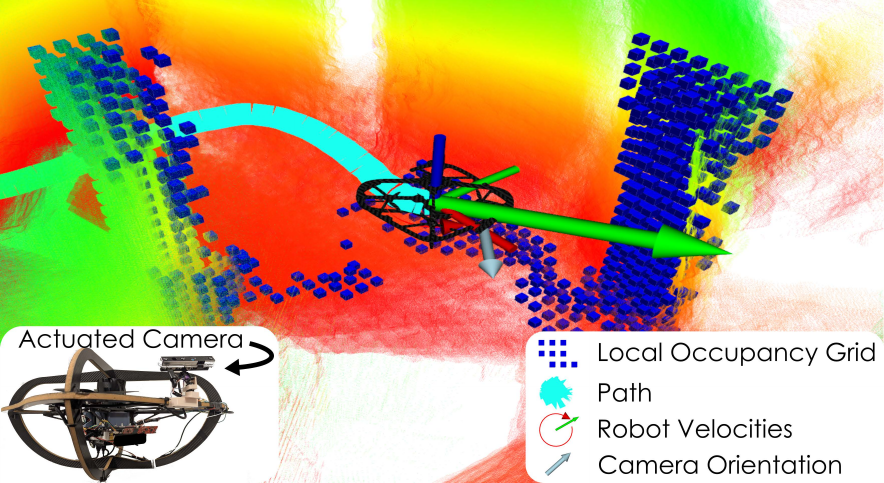}
    \vspace{-2ex}
    \caption{The quadrotor platform equipped with an actuated camera system. The local occupancy grid informs the robot about the nearby obstacle while the camera is directed to explore new regions in cluttered environments.}
    \label{fig:title_image}
    \vspace{-4ex}
\end{figure}

In parallel, \ac{rl} has emerged as a powerful paradigm for navigation in both simulated~\cite{mateus2022active} and real-world environments \cite{mirowski2016learning}. These methods learn to map high-dimensional sensory inputs, including RGB images~\cite{zhu2017target, chaplot2020object} and depth data~\cite{zhang2025learning,kulkarni2024reinforcement,lee2025quadrotor}, directly to motion commands, typically employing actor-critic or policy-gradient methods. While multi-objective \ac{rl} has been explored to jointly optimize goal-reaching and exploration objectives~\cite{chen2023multi}, most \ac{rl}-based navigation frameworks assume rigidly mounted, non-actuated, sensors with fixed orientations relative to the robot, limiting the agent's ability to actively direct its sensing apparatus toward task-relevant regions~\cite{sun2025uav}. 
Recent advances in semantic-aware and coverage-driven \ac{rl} have incorporated intrinsic rewards for exploration \cite{bartolomei2021semantic} or surface coverage \cite{malczyk2025semantically}. While \cite{malczyk2025semantically} extends the problem to 3D visual inspection, it does not consider actuated sensing and active perception-enabled navigation.

This gap between active perception theory and modern \ac{rl}-based navigation systems presents a significant opportunity. 
We argue that effective autonomy requires coupling two intertwined objectives: (i) safe, goal-directed motion, which ensures the robot reaches its target without collisions, and (ii) intrinsically-motivated informative viewpoint selection enabling the robot to actively improve scene understanding.

\subsection{Contributions}
In this work, we consider flying robots equipped with an active (actuated) camera and present a novel \ac{rl} framework that jointly optimizes for safe goal-directed motion and information gain, allowing the agent to actively decide both how to move and where to point its camera while executing its navigation task, as shown in~\Cref{fig:title_image}. Focusing on resilience, the method does not assume long-term consistent localization, and instead relies only on immediate sensor readings, locally smooth odometry and a compact yet expressive local geometry representation around the robot. The latter encodes free space and obstacles in 3D thus better allowing the agent to reason about nearby geometry for collision avoidance. Not only do we demonstrate the benefits of active perception in navigation but we also design a multi-objective reward function that combines (i) traditional navigation rewards reflecting progress to the goal, success, and collision avoidance with (ii) an information gain term that encourages the agent to maximize environment exploration and perceptual understanding without compromising either safety or navigation task completion. The method is first validated in extensive simulations,
demonstrating high target-reaching success rates and improved map completeness compared to baselines that rely on body-fixed non-actuated cameras. Moreover, we experimentally deploy the proposed trained policy on a flying robot, demonstrating that the method generalizes from simulation to physical hardware and successfully performs collision-free target-reaching and active perception in 3D environments. 
To support reproducibility, the method is open-sourced in \url{https://github.com/ntnu-arl/active-perception-RL-navigation}

The remainder of this paper is structured as follows. Sections~\ref{sec:problem_formulation} and~\ref{sec:method} describe the problem formulation and the proposed method, respectively. Evaluation is presented in \Cref{sec:results}, while conclusions are drawn in \Cref{sec:conclusion}.
\section{PROBLEM FORMULATION}
\label{sec:problem_formulation}
\begin{figure}
    \centering
    \includegraphics[width=0.98\columnwidth]{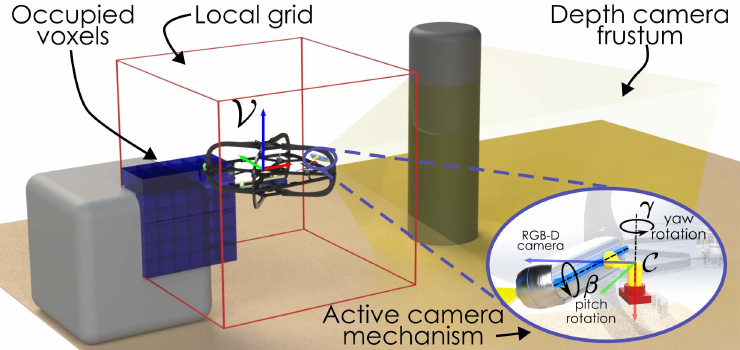}
    \vspace{-2ex}
    \caption{The quadrotor platform with its actuated RGB-D camera system.
    }
    \label{fig:hardware_image}
    \vspace{-4ex}
\end{figure}

The problem of active perception-enabled 3D navigation of aerial robots in unknown environments, as considered in this work, is that of finding a control policy allowing the robot to safely and efficiently reach a designated goal location, while simultaneously leveraging actuated sensing to enhance situational awareness. We model this as a problem of incrementally deriving a collision-free path~$\mathcal{P}_i$ to the goal location~$\mathcal{G}_i$ assuming access only to a)~the locally consistent estimate of the robot's odometry $\mathbf{s}_t$ at time $t$, b)~the current depth image $\mathbf{D}_t$, c)~an associated camera orientation $\mathbf{c}_{t}$ representing the camera pitch $\beta_t$ and yaw $\gamma_t$ of the camera frame $\mathcal{C}$ with respect to the body-fixed frame~$\mathcal{B}$, and d)~an ego-centric local occupancy grid $\mathbf{m}^o_t$ aligned with the vehicle frame~$\mathcal{V}$, as shown in~\Cref{fig:hardware_image}. The vehicle frame is yaw-aligned with the body-fixed frame, and has its \textit{x}-\textit{y} plane parallel to the inertial frame $\mathcal{I}$. The estimated robot state is defined as:
\vspace{-3ex}
\begin{align}
    \mathbf{s}_t = [ \mathbf{p}_t, \mathbf{q}_t, \mathbf{v}_t, \boldsymbol{\omega}_t],
    \label{eq:state}
\end{align}
which consists of its 3D position $\mathbf{p}_t$, orientation in a 4D vector form of the associated quaternion $\mathbf{q}_t$ expressed in $\mathcal{I}$, while the 3D linear velocity $\mathbf{v}_t$ and 3D angular velocity $\boldsymbol{\omega}_t$ are expressed in $\mathcal{B}$. The depth image~$\mathbf{D}_t$ can be obtained from an onboard RGB-D camera device (as in the studies of this work), and the local ego-centric occupancy grid $\mathbf{m}^o_t$ is built online based on the range sensor readings and locally smooth odometry. Given a 3D goal location in an unknown environment, the objective is to iteratively compute an optimal action vector that integrates navigation, obstacle avoidance, and active perception. The action vector is defined~as:
\begin{align}
    \mathbf{a}_t = [\underbrace{{\mathbf{v}_t^{r}},~{\omega}_{t,z}^{r}}_{\mathbf{a}^\textrm{nav}_t}, ~\underbrace{{\mathbf{c}_t^{r}}}_{\mathbf{a}^{\textrm{cam}}_t}],
\label{eq:action}
\vspace{-5ex}
\end{align}
where $\mathbf{v}_t^{r}\in \mathbb{R}^3$ and ${\omega}_{t,z}^{r}$ are the commanded linear velocities and yaw rate expressed in $\mathcal{V}$, and $\mathbf{c}^r_{t} = \{{\beta}_{t}^{r}, {\gamma}_{t}^{r}\}$ denotes the commanded pitch and yaw angles for the camera expressed in~$\mathcal{B}$. These can be categorized as commanded references for the low-level robot controller $\mathbf{a}^{\textrm{nav}}_t$, and orientation setpoints for the actuated camera $\mathbf{a}^{\textrm{cam}}_t$. The optimal action vector $\mathbf{a}_t$ must simultaneously satisfy a set of coupled objectives, namely: (i) navigate an unknown environment to reach the goal, (ii) maintain collision-free flight in the presence of obstacles, and (iii) actively gain awareness of the environment through joint optimization of robot motion and camera orientation. The last objective enhances performance without compromising the navigation and safety goals.

\section{METHOD}
\label{sec:method}
\begin{figure*}[t]
    \centering
    \includegraphics[width=0.99\textwidth]{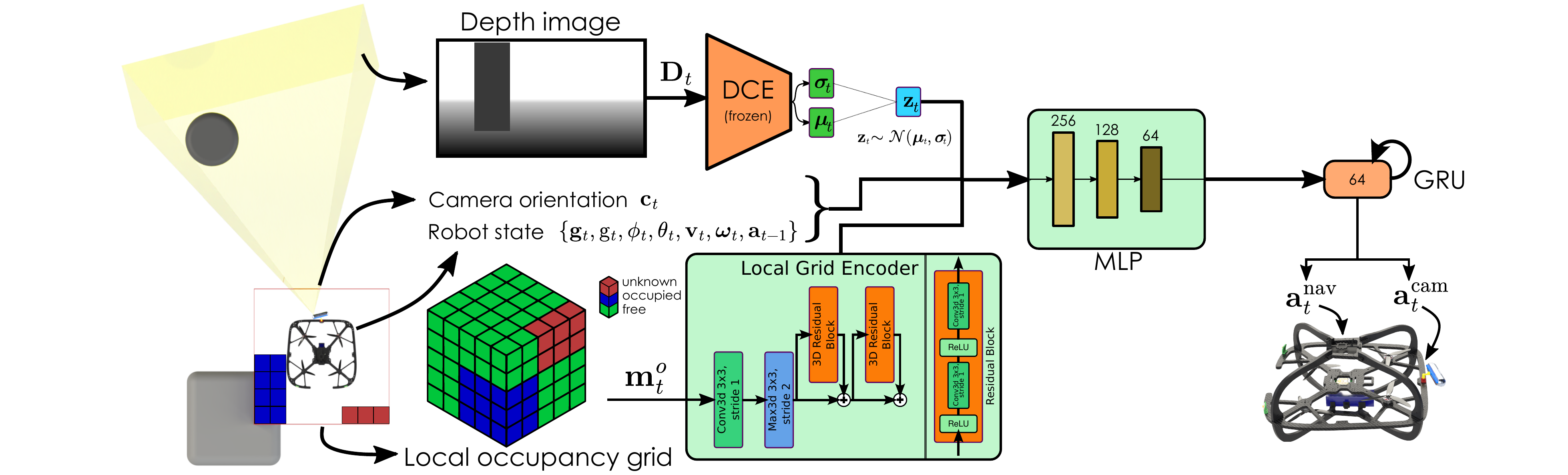}
    \vspace{-2ex}
    \caption{The proposed network architecture for safe navigation with active perception. The network processes depth images as well as the local occupancy grids through dedicated encoder blocks, integrates robot state and camera orientation via MLP and GRU modules, and commands the actions for the robot.}
    \label{fig:network_architecture}
    \vspace{-4ex}
\end{figure*}

We formulate the active perception-enabled collision-free navigation as a reinforcement learning task. We define the state space $\mathcal{S}$ as the set of all possible agent and environment states with $\mathfrak{s}_t \in \mathcal{S}$ at discrete time $t$. The action space is denoted as $\mathcal{A}$ with $\mathbf{a}_t\in\mathcal{A}$. We denote the observation space as $\mathcal{O}$ with each agent-received observation denoted as $\mathfrak{o}_t \in \mathcal{O}$. Finally, $\mathcal{R}$ represents the reward function. Subsequently, we define how we construct and derive each of these quantities of the \ac{rl} problem toward safe navigation with active perception.

\subsection{Ego-centric Local Occupancy Grid}
\label{sec:occupancy}
To enable safe navigation, our approach utilizes a compact, local representation of the robot's immediate surroundings. Unlike conventional planning methods~\cite{oleynikova2016continuous} that plan on a global occupancy map $\mathbf{M}^o_t$ susceptible to localization drift with time, our \ac{rl} framework operates on a local, ego-centric occupancy grid $\mathbf{m}^o_t$. This design choice enhances generalization across diverse environments and significantly improves computational efficiency.

A dense map expressed in $\mathcal{I}$ continuously integrates depth measurements, using the (possibly drifting) odometry estimates. From this, the local occupancy grid $\mathbf{m}^o_t$ is extracted around the vicinity of the robot at time $t$. To construct this representation, depth sensor measurements are first transformed from the camera frame to the vehicle frame. For each depth image pixel, rays are cast up to a maximum distance $d_m$ (here $3$ m), marking voxels containing points corresponding to obstacles as occupied, while the voxels up to this point are marked as free. Voxels that have never been observed remain marked as unknown. This mapping process is performed incrementally in a local region around the robot, temporally accumulating information as the robot moves. Temporal integration helps mitigate sensor noise and provides a more complete representation of the local environment than instantaneous depth measurements alone. Simultaneously, as this map is built only for a short range around the robot and is updated iteratively, it does not require long-term consistent odometry. 
Overall, this local representation provides the \ac{rl} agent with the essential additional information needed for improved collision avoidance as compared to methods that only assume access to instantaneous camera data~\cite{kulkarni2024reinforcement,nguyen2024uncertainty,loquercio2021learning}. It enables more robust policies that proactively avoid collisions with obstacles located outside the sensor frustum. The local ego-centric grid is illustrated in~\Cref{fig:network_architecture}.

\subsection{Policy Learning}
\paragraph{State and Observation Space}
The underlying state $\mathfrak{s}_t \in \mathcal{S}$ encodes the robot’s 6-DoF pose, target location, and the complete geometry of the surrounding environment. Since the state is not directly observable, the agent instead receives an observation $\mathfrak{o}_t \in \mathcal{O}$, defined as:
\begin{align}
    \mathfrak{o}_t = \{{\mathbf{g}_t}, \mathrm{g}_t, \phi_t, \theta_t ,\mathbf{v}_t, \boldsymbol{\omega}_t, \mathbf{c}_t, \mathbf{a}_{t-1}, \mathbf{z}_{t}, \mathbf{m}^o_t\},
\label{eq:observation}
\end{align}
where $\mathbf{g}_t \in \mathbb{R}^3$ is a unit vector to the target goal location expressed in $\mathcal{V}$ and the corresponding distance $\mathrm{g}_t \in \mathbb{R}$. The robot's attitude is represented through pitch $\phi_t$ and roll $\theta_t$ angles, expressed in~$\mathcal{V}$. In addition, linear velocity~$\mathbf{v}~\in~\mathbb{R}^3$ and angular velocity $\boldsymbol{\omega} \in \mathbb{R}^3$, expressed in $\mathcal{B}$ are included to inform the policy about system dynamics. The actuated camera orientation is observed through its pitch and yaw angles $\mathbf{c}_t = \{\beta_t, \gamma_t\}$, while the previous control actions $\mathbf{a}_{t-1} \in \mathbb{R}^6$ are appended to the state as well. The high-dimensional depth image $\mathbf{D}_t$ is compressed into latent embeddings $\mathbf{z}_{t}$ produced using the \ac{dce}~\cite{kulkarni2023task}, and appended to the observation vector. \ac{dce} focuses on retaining collision information, while aggressively compressing the input depth image. Beyond vectorized features, the agent also receives the local occupancy grid $\mathbf{m}^o_t$ represented in $\mathcal{V}$, enhancing spatial context for obstacle avoidance.
\paragraph{Action Space}
The action space jointly encompasses navigation and active perception. At each control step, the policy outputs commands for the aerial robot's motion (${\mathbf{a}^{\mathrm{nav}}_t}$) and desired orientation for the actuated camera (${\mathbf{a}^{\mathrm{cam}}_t}$). The first, $\mathbf{a}_t^{\mathrm{nav}} \in \mathbb{R}^4$, is parameterized to represent the commanded linear velocities $\mathbf{v}_t^{r}$ and the yaw rate ${\omega}_{t,z}^{r}$. 
We thus allow the agent to fully utilize the range of possible motions and efficiently explore the environment without constraining the commands to strictly lie within the \ac{fov} of the depth camera~\cite{kulkarni2024reinforcement}.
The second, $\mathbf{a}_t^{\mathrm{cam}}:= {\mathbf{c}_t^{r}}$, specifies the desired pitch and yaw of the actuated camera. These actions are bounded within hardware limits $\beta^{r}_t \in [-\beta_{\max},\beta_{\max}]$, $\gamma^{r}_t \in [-\gamma_{\max},\gamma_{\max}]$, ensuring feasibility with respect to the servo actuation. The overall action vector is thus defined as:
\begin{equation}
    \mathbf{a}_t = \big[{\mathbf{a}^{\mathrm{nav}}_t}, \, {\mathbf{a}^{\mathrm{cam}}_t}\big] \in \mathbb{R}^6 ,
\end{equation}
allowing the agent to simultaneously control the robot's motion and actively orient the camera. This unified action space enables the policy to ensure collision-free navigation, and at the same time efficient actuation of the camera to both maximize information for collision-free navigation and exploration of the environment.

\paragraph{Reward Design}
The agent receives a reward $\mathcal{R}(\mathfrak{s}_t, \mathbf{a}_t)$ for each state transition, based on which it learns a policy $\pi$ that maps observations and belief states to actions: $\mathbf{a}_t = \pi(\mathfrak{o}_{t}, \mathfrak{b}_{t})$, where $\mathfrak{b}_{t}$ represents the agent's belief distribution over possible environment states given its observation history. During training, a global occupancy grid of the environment $\mathbf{M}^o_t$ serves as privileged knowledge to: (i) compute reward signals, (ii) assess safety violations with ground truth obstacle locations, and (iii) determine which voxels in the global environment have been observed by the agent's sensors. This enables accurate reward computation and proper supervision during the learning process. However, this global information is deliberately withheld from the agent's policy, which must operate solely on the local observations $\mathfrak{o}_t$ that provide only partial, noisy measurements of nearby obstacles and free space. This training paradigm ensures that the learned policy remains deployable in real-world scenarios where global environmental knowledge is unavailable.
The reward function is defined as:
\begin{align}
\mathcal{R}(\mathfrak{s}_t, \mathbf{a}_t) = r_t + l_t + n_t +  p_t,
\end{align}
where each term serves a distinct purpose. The term $r_t$ rewards the agent for getting closer to the target location based on the current distance and is defined as:
\begin{align}
r_t = \lambda_d (d_{t-1} - d_t) + \lambda_e (e^{-\alpha d_t^2}),
\end{align}
where $d_t$ is the Euclidean distance between the robot's current position and the target at time $t$, and $\lambda_d, \lambda_e , \alpha \in \mathbb{R}_{+}$ are scaling factors. The exponential term encourages the agent to make faster progress as it gets closer to the goal. The term $l_t$ penalizes the agent for jerky or large movements of both the robot and the camera, promoting smooth motions. It is defined as follows:
\begin{align}
l_t = -\lambda_a||\mathbf{a}_{t} - \mathbf{a}_{t-1}||,
\end{align}
where $\lambda_a \in \mathbb{R}^6_{+}$ is a scaling factor. The term $n_t$ encourages the agent to actively explore and discover new information. This intrinsic reward is computed based on privileged information from the global occupancy grid (distinct from the local occupancy grid $\mathbf{m}^o_t$ used by the policy), and it is proportional to the number of voxels whose state transitions from \textit{unknown} to either \textit{free} or \textit{occupied} between time steps $t-1$ and $t$. This incentivizes the agent to re-orient its camera towards previously unobserved regions, thereby improving its situational awareness. $n_t$ is defined as:
\begin{align}
\scriptsize
n_t = \lambda_{\mathcal{G}}\sum_{i \in \mathbf{M}^o_t}\mathbb{I}[ \text{state}(i)_t \neq \text{unknown} \land \text{state}(i)_{t-1} = \text{unknown}],
\vspace{-2ex}
\end{align}
\normalsize
where $\lambda_{\mathcal{G}}\in\mathbb{R}_{+}$ is a scaling factor and $\mathbb{I}[\cdot]$ is the indicator function. The sum iterates over all voxels in the global map, and the indicator function evaluates to $1$ if a voxel transitions from an \textit{unknown} state to a known state (\textit{free} or \textit{occupied}) and 0 otherwise. Note that this term is not used by default unless explicitly specified. Lastly, the distance penalty $p_t$ is designed to prevent collisions. It is proportional to the number of occupied and unknown voxels within a critical collision distance $d_{coll}>0$ from the robot. $d_{coll}$ reflects the robot's physical dimensions. The penalty $p_t$ is calculated as:
\vspace{-1ex}
\begin{align}
\scriptsize
p_t = -\lambda_p\sum_{i \in \text{sphere}(d_{coll})} \mathbb{I}[\text{voxel}_i \in \{\text{occupied, unknown}\}],
\vspace{-2ex}
\end{align}
\normalsize
where $\lambda_p\in \mathbb{R}_{+}$ is a scaling factor. A high value of $p_t$ indicates the robot is in close proximity to obstacles, encouraging the agent to steer away.

\subsection{Implementation}
\paragraph{Actuated Camera}
We model the response of the camera actuators with first-order dynamics matching the real servos. Given commanded angles $\beta^{r}$, $\gamma^{r}$, the joint dynamics are:
\vspace{-2ex}
\begin{align}
\dot{\beta}_t &= \frac{1}{\tau_\beta}\operatorname{sat}_{[-\beta_{\max},\,\beta_{\max}]}\left(\beta^{r}_t - \beta_t\right),
\label{eq:first_order_theta}\\
\dot{\gamma}_t &= \frac{1}{\tau_\gamma}\operatorname{sat}_{[-\gamma_{\max},\,\gamma_{\max}]}\left(\gamma^{r}_t - \gamma_t\right),
\label{eq:first_order_psi}
\end{align}
\normalsize
where $\tau_\beta,\tau_\gamma$ are time constants identified from the physical servos, and $\operatorname{sat}_{[a,b]}(\cdot)$ represents the saturation function that clips values to the interval $[a,b]$.
For a first-order system, the $10$--$90\%$ rise time is $t_r \approx 2.2\,\tau$, which we utilize to calibrate the time constant $\tau$ from step-response measurements.  The discrete-time formulation implemented in simulation employs a forward-Euler integration scheme:
\begin{align}
\beta_{t+1} &= \beta_{t} + \frac{\Delta t}{\tau_\beta}\!\left(\tilde{\beta}_t^{r} - \beta_{t}\right), \\
\gamma_{t+1} &= \gamma_{t} + \frac{\Delta t}{\tau_\gamma}\!\left(\tilde{\gamma}_t^{r} - \gamma_{t}\right),
\end{align}
\normalsize
with saturated commands $\tilde{\beta}^{r}_t=\operatorname{sat}_{[-\beta_{\max},\,\beta_{\max}]}(\beta_t^{r})$ and $\tilde{\gamma}^{r}_t=\operatorname{sat}_{[-\gamma_{\max},\,\gamma_{\max}]}(\gamma_t^{r})$.

\paragraph{Network architecture}
We employ the \ac{appo} algorithm from \cite{petrenko2020sample} to train a deep neural network policy that enables collision-free robot navigation exploiting active perception. Similar to~\cite{malczyk2025semantically}, building upon the 2D model architecture, we extend the approach to 3D by adapting a ResNet-based encoder with hyperparameters from~\cite{espeholt2018impala} to effectively process spatial information from the local 3D occupancy grid inputs $\mathbf{m}^o$. 
Ego-centric occupancy maps of size $n\times n\times n$, where $n=21$, are discretized using a grid resolution of $r_V=$~\SI{0.1}{m}, chosen to provide sufficient spatial local detail with the collision distance $d_{coll}$ set to \SI{0.4}{\meter} respecting the real-world robot size. In turn, this also implies that the local occupancy sizes $2.1\times 2.1\times 2.1$~\SI{}{m}$^3$. The odd number of cells ensures that the robot is always centered symmetrically at the middle voxel of the local occupancy grid. The ResNet encoder learns compressed latent representations of spatial features from these local occupancy grids, enabling the policy to efficiently acquire collision avoidance behaviors for objects even when they are not present anymore within the depth sensor~\ac{fov}.

The compressed representations from the pre-trained and frozen \ac{dce} are combined with the representations from the grid encoder and the robot and camera states. These are then fed into a \ac{mlp}. This \ac{mlp} consists of three fully connected layers of size $256$, $128$ and $64$ neurons each, with an ELU activation layer, followed by a \ac{gru} block with a hidden size of~$64$. Given an observation vector~$\mathfrak{o}_t$, the policy outputs a $6$-dimensional action command $\mathbf{a}_t$. Linear velocity commands are scaled between $\pm$~\SI{1}{\meter/\s}, yaw rate is scaled between $\pm$~\SI{1}{\radian/\s}. Similarly, the commands for camera orientation are scaled between their maximum ranges $\pm$~$\beta_{\textrm{max}}$ and $\pm$~$\gamma_{\textrm{max}}$ for pitch and yaw respectively. The commands are then sent to the low-level velocity controller onboard the robot and the actuated camera, as shown in \Cref{fig:network_architecture}. Simultaneously, with $\beta_{\max} = \gamma_{\max} = \frac{\pi}{2}$ \SI{}{\radian}, the camera angle command for pitch and yaw is fed as reference to the servo controllers.
\subsection{Training Environment}
For training, we utilize the Aerial Gym Simulator \cite{kulkarni2025aerial}, which provides the environment and the interfaces to train our deep reinforcement learning policy to navigate within various environments. The simulator offers capabilities for massively parallelized simulation of aerial robots with exteroceptive sensors. The simulated flying platform uses the velocity controller from \cite{lee2010geometric} and is equipped with a depth camera with horizontal and vertical \ac{fov} of $\{86,57\}^\circ$, and a sensing range of \SI{0.2}{} to \SI{10}{\meter}. 

\begin{figure}
    \centering
    \includegraphics[width=0.98\columnwidth]{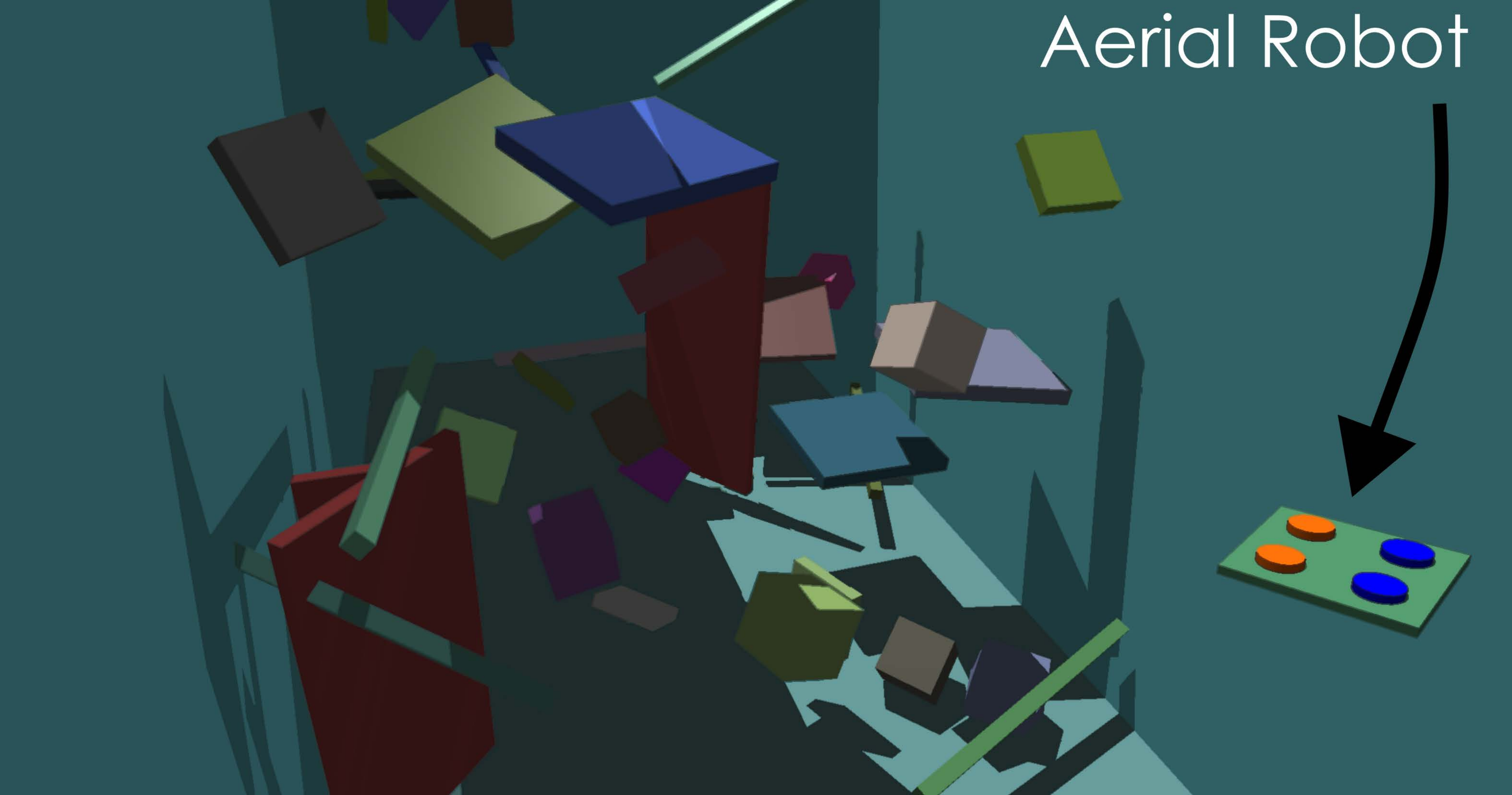}
    \vspace{-2ex}
    \caption{Aerial Gym training environment with randomized primitive obstacles ensuring diverse scenarios and sim2real transferability.
    }
    \label{fig:aerialgym}
    \vspace{-5ex}
\end{figure}

We generate the environments within the simulator, consisting of corridor-like scenes containing randomly placed static obstacles of primitive shapes and different sizes, as shown in~\Cref{fig:aerialgym}. The utilization of primitive geometric objects relates to the goal of generalizability of navigation, as these fundamental shapes provide diverse geometric challenges without introducing domain-specific complexities. The corridor-shaped environments have dimensions $L~\times~W~\times~H$ within the set $[10, 12] \times [5, 8] \times [4, 6]~\SI{}{\meter}$. The start and goal locations are randomly sampled at opposite ends of the environment for each episode. The agent's initial yaw angle is uniformly sampled from $\mathcal{U}(\frac{-\pi}{2}, \frac{\pi}{2})$ to avoid bias toward straight-line trajectories. Each episode spans $\SI{10}{\s}$ considering environment size and maximum robot speeds.

To robustify the network performance against real-world uncertainty, we introduce multiple sources of randomization and noise. Let $\mathcal{N},~\mathcal{U}$ represent the normal and uniform distributions, respectively. Disturbance wrenches $\mathbf{w} \sim \mathcal{N}(\mathbf{0}, \sigma_w^2\mathbf{I})$ with $\sigma_w = \SI{5}{\newton}$ are applied to the simulated platform, while observations from~\Cref{eq:observation} are perturbed by noise $\boldsymbol{\epsilon}_s \sim \mathcal{U}(-\delta_s, \delta_s)$ with $\delta_s = \SI{0.1}{\meter}$ for position and $\delta_s = \SI{0.05}{\meter\per\second}$ for velocity components. Camera sensor position and orientation are perturbed by $\boldsymbol{\epsilon}_p \sim \mathcal{U}(-\SI{5}{\centi\meter}, \SI{5}{\centi\meter})$ and $\boldsymbol{\epsilon}_\theta \sim \mathcal{U}(-\SI{5}{\degree}, \SI{5}{\degree})$, respectively. Additionally, velocity controller parameters are randomized as $\tau_{nominal} \pm 12\%$ to vary step-response characteristics, and depth images are corrupted with Gaussian noise $\boldsymbol{\epsilon}_d \sim \mathcal{N}(0, \sigma_d^2(z))$, combined with random pixel dropout at probability $p = 0.01$ mimicking realistic depth sensor characteristics~\cite{khoshelham2012accuracy}.
Finally, we set the image capture and control rate to $\SI{10}{Hz}$, while the physics simulation occurs at $\SI{100}{Hz}$. The policy trains in approximately two hours on an NVIDIA RTX A6000. 
\section{EVALUATION STUDIES}
\label{sec:results}
We demonstrate the effectiveness of our active perception-enabled navigation policy through a comprehensive evaluation spanning both high-fidelity simulations and real-world robotic deployments. Our experiments validate the method's performance across diverse environmental configurations, revealing its robust generalization capabilities and successful bridging of the sim2real gap.

\begin{table*}[t]
\centering
\caption{Navigation and environment exploration performance comparison between static and active camera configurations. \\ A set of active camera policies are investigated with the last further focusing on scene exploration.}
\vspace{-2ex}
\resizebox{\textwidth}{!}{
\begin{tabular}{@{}lccccccccccccc@{}}

\toprule
\multicolumn{1}{c|}{\# obstacles}                                       & \multicolumn{3}{c}{0}                & \multicolumn{3}{c|}{10}                & \multicolumn{3}{c|}{20}                 & \multicolumn{4}{c}{30} \\
\multicolumn{1}{l|}{~}                                      & Success & Timeout & \multicolumn{1}{c|}{Crash} & Success & Timeout & \multicolumn{1}{c|}{Crash}& Success & Timeout & \multicolumn{1}{c|}{Crash} & Success & Timeout & \multicolumn{1}{c}{Crash} & \textit{Exploration}  \\ \midrule

\multicolumn{1}{l|}{Static}                                      & 99.3\% & 0.1\% & \multicolumn{1}{c|}{0.6\%} & 82.4\% & 7.2\% & \multicolumn{1}{c|}{10.4\%} & 70.7\% & 4.8\%  & \multicolumn{1}{c|}{24.5\%} & 65.3\%   & 2.3\%   & \multicolumn{1}{c}{32.4\%} & 26.5\% \\
\multicolumn{1}{l|}{Static+FOV}                & 99.4\% & 0.6\% & \multicolumn{1}{c|}{0.0\%} & 90.4\% & 2.2\% & \multicolumn{1}{c|}{6.6\%}  & 79.6\% & 3.2\%  & \multicolumn{1}{c|}{17.2\%} & 71.5\%   & 8.9\%   & \multicolumn{1}{c}{19.6\%} & 24.0\% \\
\multicolumn{1}{l|}{Static+Grid}                      & 99.6\% & 0.4\% & \multicolumn{1}{c|}{0.0\%} & 91.3\% & 6.7\% & \multicolumn{1}{c|}{2.0\%}  & 82.4\% & 14.7\% & \multicolumn{1}{c|}{2.9\%}  & 85.5\%   & 9.7\%  & \multicolumn{1}{c}{4.8\%}  & 29.6\% \\
\multicolumn{1}{l|}{Static+Grid+FOV} & 99.8\% & 0.2\% & \multicolumn{1}{c|}{0.0\%} & 91.4\% & 4.4\% & \multicolumn{1}{c|}{4.2\%}  & 87.2\% & 5.7\%  & \multicolumn{1}{c|}{7.1\%}  & 86.0\%   & 8.9\%   & \multicolumn{1}{c}{5.1\%}  & 28.3\% \\
\midrule

\multicolumn{1}{l|}{Active}                                      & 99.7\% & 0.2\% & \multicolumn{1}{c|}{0.1\%} & 92.5\% & 1.7\% & \multicolumn{1}{c|}{5.8\%}  & 86.4\% & 0.4\%  & \multicolumn{1}{c|}{14.0\%} & 83.2\%   & 0.5\%   & \multicolumn{1}{c}{16.3\%} & 41.5\% \\
\multicolumn{1}{l|}{Active+FOV}                                      & 99.9\% & 0.1\% & \multicolumn{1}{c|}{0.0\%} & 97.8\% & 1.7\%& \multicolumn{1}{c|}{0.5\%}  & 95.9\% & 2.1\% & \multicolumn{1}{c|}{2.0\%} & 94.9\%   & 2.4\%   & \multicolumn{1}{c}{2.7\%}   & 41.2\% \\
\multicolumn{1}{l|}{Active+Grid}                                      & 99.9\% & 0.0\% & \multicolumn{1}{c|}{0.1\%} & 98.4\% & 0.9\%& \multicolumn{1}{c|}{0.7\%}  & 96.2\% & 1.6\% & \multicolumn{1}{c|}{2.2\%} & {95.4\%}   & 2.0\%   & \multicolumn{1}{c}{2.6\%}   & {43.4}\% \\
\multicolumn{1}{l|}{Active+Grid+FOV}                                      & 99.8\% & 0.2\% & \multicolumn{1}{c|}{0.0\%} & 98.2\% & 1.5\%& \multicolumn{1}{c|}{0.3\%}  & 95.9\% & 2.0\% & \multicolumn{1}{c|}{2.1\%} & {95\%}   & 2.5\%   & \multicolumn{1}{c}{2.5\%}   & 42.6\% \\

\multicolumn{1}{l|}{$\triangleright$ Active+Grid+$n_t$}                      & 99.9\% & 0.1\% & \multicolumn{1}{c|}{0.0\%} & 97.4\% & 1.5\% & \multicolumn{1}{c|}{1.1\%}  & 96.0\% & 1.7\%  & \multicolumn{1}{c|}{2.3\%}  & 94.3\%   & 2.8\%   & \multicolumn{1}{c}{2.9\%}  & {63.4\%}\\
\bottomrule
\end{tabular}
}
\vspace{-5ex}
\label{tab:navigation}
\end{table*}

\subsection{Ablation Studies}
To evaluate the individual contributions of our method's key components, we conduct comprehensive ablation studies in the Aerial Gym Simulator to isolate the effects of active camera control and local grid representation. These studies examine nine different configurations across environments of varying obstacle density, allowing us to quantify the performance gains attributable to each architectural choice. The evaluated approaches include: (i) \textit{Static} baseline using a fixed camera orientation where the policy can plan navigation actions in any direction, including outside the current camera \ac{fov} (creating an intentional challenge for perception-action coordination), (ii) \textit{Static+FOV}, proposed by~\cite{kulkarni2024reinforcement}, where navigation actions are restricted to move only in directions currently visible within the camera's \ac{fov} (i.e., the agent cannot command velocities outside the sensor frustum), (iii) \textit{Static+Grid} incorporating local occupancy grid representation while maintaining fixed camera orientation, (iv) \textit{Static+Grid+FOV} combining both local grid input and \ac{fov}-constrained actions with static camera, (v) \textit{Active} enabling dynamic camera control, (vi) \textit{Active+FOV} introducing the action constraints within the instantaneous camera \ac{fov} coupled with the ability to actively reorient the camera, (vii) \textit{Active+Grid} integrating active camera control with local occupancy grid representation, (viii) \textit{Active+Grid+FOV} which restricts the previous to move only in directions currently visible within the actuated camera's \ac{fov}, and finally (ix) \textit{Active+Grid+$n_t$} that augments \textit{Active+Grid} with exploratory sensing reward.

The results, presented in~\Cref{tab:navigation}, demonstrate the significant impact of both active perception and local grid representation on navigation performance across environments of varying complexity. Note that success is defined as reaching the target within \SI{1}{\meter} range, crash indicates collision with an obstacle, and a timeout is induced after \SI{10}{\second} without success or collision. In obstacle-free environments (corridor with $0$ floating obstacles), all approaches achieve near-perfect success rates ($\geq$99.3\%), indicating that the fundamental navigation capability is well-established across all configurations. However, as environmental complexity increases, the performance differences become pronounced. The baseline static camera approach shows substantial degradation with increasing obstacle density, achieving only $65.3$\% success in the most complex scenario (30 obstacles) with a concerning~$32.4$\% crash rate. The approach incorporating actions constrained within the current camera \ac{fov}, provides modest improvements, increasing success rates to $71.5$\% in dense environments. The introduction of local grid representation yields the most substantial gains, significantly reducing crash rates to $4.8$\%, while boosting success rates. Introduction of the local occupancy grid significantly outweights the effect of constraining actions to the sensor \ac{fov} for success rates.

Naturally, the introduction of the active camera with either the local grid or motion constrained with the \ac{fov} enables the robot to achieve significantly lower crash rates~($\leq$2.7\%) across all complexity levels, while maintaining higher success rates. Interestingly, only the active camera approach without the local grid representation has a higher crash rate than the static camera methods using the local occupancy grid. The reduction of crash rates with the introduction of the local occupancy grid is consistent across all ablations, highlighting its necessity. Finally, the \textit{Active+Grid+$n_t$} approach achieves comparable navigation success rates of $97.4$\%, $96.0$\%, and $94.3$\% for $10$, $20$, and $30$ obstacles respectively, demonstrating that active perception, when combined with effective spatial representation, enables robust navigation performance that scales well with environmental complexity.
\begin{figure}
    \centering
    \includegraphics[width=0.97\columnwidth]{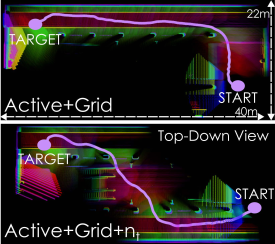}
    \vspace{-2ex}
    \caption{Gazebo train station navigation experiment. Top-down view comparing trajectories of two methods: one without $n_t$ incorporation and one with $n_t$ The latter demonstrates improved spatial awareness, scaning more of the enviroment while navigating towards the goal.
    }
    \label{fig:gazebo}
    \vspace{-4.5ex}
\end{figure}
Beyond navigation success rates, the ablation studies reveal substantial variations in spatial exploration efficiency, as demonstrated in right most column of~\Cref{tab:navigation}. We quantify the exploration performance by measuring the percentage of environment volume discovered during the navigation task, calculated as the ratio of voxels that transition from unknown to either free or occupied states relative to the total environment volume, based on the privileged ground-truth occupancy grid. Evaluations are done in an environment with $30$ obstacles. Static camera configurations achieve relatively limited environmental awareness, with exploration ranging from $24.0$\% to $29.6$\% across different input variations. However, the active camera approaches demonstrate substantially superior environmental exploration capabilities. The basic active camera configuration can explore up to $43.4$\%, highlighting the importance of active perception for environment understanding. The introduction of local grid representation does not significantly impact this metric, highlighting its dedicated contribution towards improving robot safety, with limited effect towards exploration behavior. Approach {(ix)~\textit{Active+Grid+$n_t$}} achieves the highest exploration of $63.4$\%, representing a $139$\% improvement over the baseline static approach. This shows that active perception not only improves both navigation safety and success rates but also significantly enhances the robot's ability to gather environmental information during task execution, supporting better environmental awareness and potentially enabling more informed decision-making in complex scenarios.
\begin{figure*}
    \centering
    \includegraphics[width=0.99\textwidth]{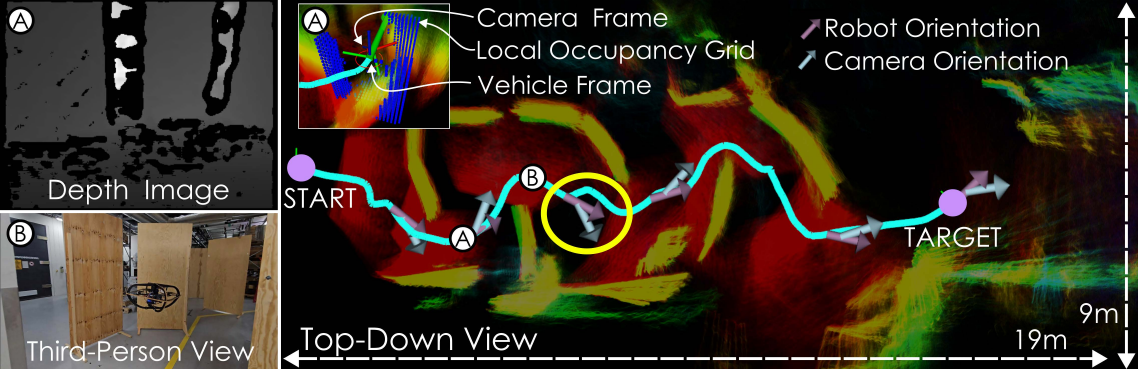}
    \vspace{-2ex}
    \caption{Top-down view of navigation through a cluttered corridor. The cyan trajectory and point cloud show the robot's path and perception. Gray and purple arrows indicate camera orientations diverging from robot heading for enhanced spatial awareness. Network inputs (depth image, local occupancy grid) are shown at time A; a third-person mission view at time B.
    }
    \label{fig:missionCorridor}
    \vspace{-4ex}
\end{figure*}
\subsection{Simulation Studies}

To further investigate the impact of the intrinsic exploration reward $n_t$, we conduct simulation experiments in Gazebo using a train station environment~(\Cref{fig:gazebo}). The quadrotor is tasked with reaching a target location while avoiding collisions. We compare two variants of the policy, namely \textit{Active+Grid} and \textit{Active+Grid+$n_t$}. The latter is augmented with the term $n_t$, which encourages the agent to actively explore and discover new spatial information. For a controlled comparison, we establish a standardized protocol where the robot consistently starts at the same location (next to the bottom track in the top-down view) and navigates to an identical target location approximately~\SI{42.5}{\meter} away on the opposite end of the train station platform next to the other track. This setup ensures that performance differences are attributable to the reward formulation rather than environmental variations and show that the approach is not limited to the straight corridor-like environments, that the method was trained in. In both cases, no prior information for the environment is provided to the agent.

The results, summarized in~\Cref{tab:navigation}, confirmed that the inclusion of the reward $n_t$ does not compromise navigation performance, as both policy variants achieve comparable success rates in reaching the target. Simultaneously, the $n_t$-augmented policy yields a remarkable improvement in exploration of the environment. In the Gazebo train station scenario (\Cref{fig:gazebo}), the agent with active camera control and $n_t$ discovers approximately $61\%$ of the environment, compared to around $47.5\%$ for the variant without $n_t$ across $5$ runs with identical start and target locations. These findings demonstrate that the intrinsic reward promotes richer scene understanding and broader spatial exploration without reducing goal-reaching capability, thereby validating $n_t$ as a valuable signal during training.

\subsection{Real-world Evaluations}
A custom-built quadrotor platform equipped with an \ac{imu}, a radar sensor for odometry estimation, and an actuated RGB-D camera system, as shown in \Cref{fig:hardware_image} is considered in this work. The actuated perception system consists of an Intel RealSense D455 camera mounted on a two-axis actuation mechanism at the edge of the robot frame. Two servo motors arranged in a pan-tilt configuration, enable independent joint position control of the camera’s pitch and yaw. Each servo motor is equipped with an integrated potentiometer, which provides real-time feedback on its joint position. This provides with an accurate closed-loop position control of the camera, ensuring that the desired orientation is precisely achieved. The mechanism allows for a rotation of $\pm \SI{45}{\degree}$ along yaw and $\pm \SI{60}{\degree}$ along pitch axis. 
The RGB-D camera delivers synchronized color and depth streams at up to \SI{10}{\Hz}, facilitating tasks such as scene understanding and obstacle avoidance with spatial awareness. The platform is powered by an onboard NVIDIA Jetson Orin NX \SI{16}{\giga\byte} module, which provides the necessary compute for running high-level navigation modules. Low-level attitude stabilization and motor control are handled by a PX4 flight controller. Communication between the high-level stack and the flight controller is achieved through ROS middleware. The \textit{Active+Grid+$n_t$} policy is chosen for these experiments as it offers the simultaneous benefits of both safe navigation and exploration. The policy is executed onboard the robot at \SI{10}{\Hz}.

\paragraph{Maneuvering in a cluttered corridor}
To validate the practical applicability of our approach and demonstrate the sim2real transfer capabilities, we deploy the \textit{Active+Grid+$n_t$} method in a cluttered indoor corridor environment, as depicted in~\Cref{fig:missionCorridor}. The experimental setup consists of a narrow corridor (\SI{15}{\meter} long, \SI{3.0}{\meter} wide) populated with obstacles and structural elements that create a challenging navigation scenario requiring precise maneuvering and collision avoidance. The policy demonstrates remarkable performance in successfully navigating tight spaces and around obstacles, with the camera looking around while traversing the environment. As highlighted by the yellow circle in~\Cref{fig:missionCorridor}, the agent proactively directs the camera toward unexplored regions to acquire comprehensive spatial understanding of its surroundings, independent of the current navigation heading. 
\paragraph{Spatial awareness in T-shaped corridor}
\begin{figure*}
    \centering
    \includegraphics[width=0.99\textwidth]{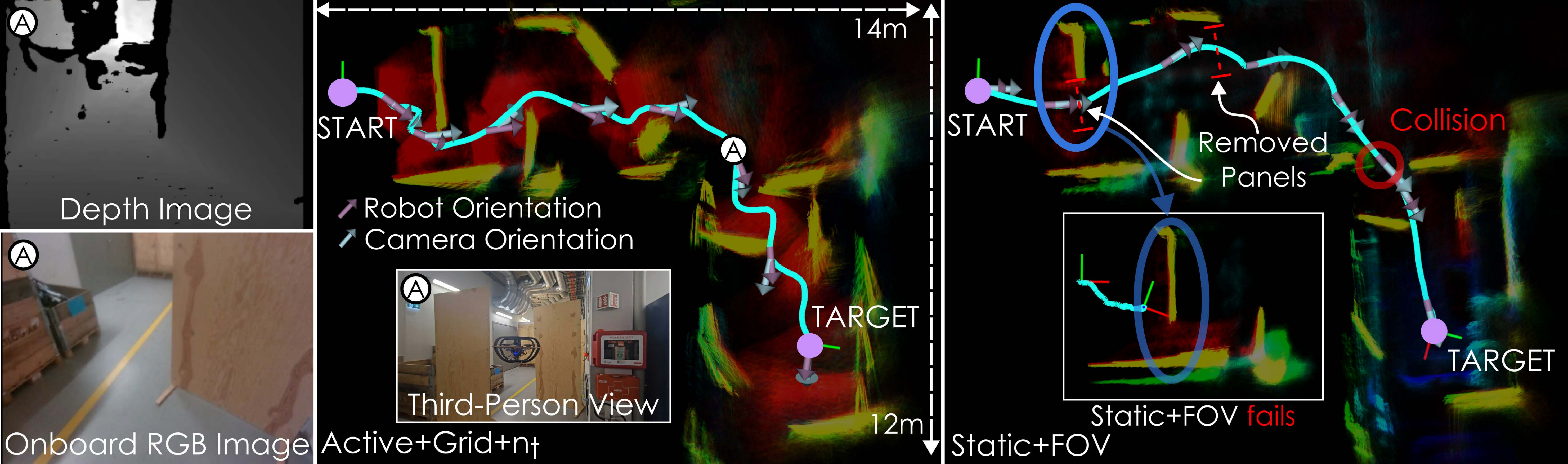}
    \vspace{-2ex}
    \caption{Navigation missions in a T-shaped corridor comparing (\textit{Active+Grid+$n_t$}) with \textit{Static+FOV}~\cite{kulkarni2024reinforcement}. The trajectories highlight the differences in navigation performance. 
    For the \textit{Static+FOV} method, the environment was modified to be less cluttered in order to enable successful navigation. Red circle indicates collision point with the environment.
    }
    \label{fig:missionLCorridor}
    \vspace{-4ex}
\end{figure*}
To demonstrate the benefits of active perception in another real-world deployment, we conduct an experiment in a T-shaped corridor environment~(\Cref{fig:missionLCorridor}). This scenario requires the quadrotor to make navigation decisions at a T-intersection to reach the target location. The active perception policy successfully adapts by reorienting the actuated RGB-D sensor to scan the lateral branch of the corridor, thereby acquiring important spatial information before committing to a turning maneuver. As a result, the robot avoids premature or incorrect navigation decisions while maintaining progress toward the target, successfully reaching it. We compare the \textit{Active+Grid+$n_t$} policy  against the \textit{Static+FOV} method~\cite{kulkarni2024reinforcement} in the same environment. As shown in the right part of~\Cref{fig:missionLCorridor}, the \textit{Static+FOV} method fails to navigate in the environment and cannot progress toward the target location, specifically getting stuck at narrow regions. To enable a comparison with~\cite{kulkarni2024reinforcement}, we are compelled to modify the environment by removing two panels. Even in this less cluttered configuration, the static camera method collides with the environment mid-trajectory (marked with the red circle), though it ultimately reaches the target location.
This experiment highlights how active perception enables the system to build a richer situational awareness in challenging environments, which is not achievable when the camera remains fixed in a forward-facing orientation.
\section{CONCLUSION}
\label{sec:conclusion}
This paper introduced a novel \ac{rl} framework for autonomous aerial navigation that integrates active perception into the control policy. We addressed the challenge of coupling motion planning with information-driven viewpoint selection by designing a multi-objective reward function that combines navigation goals with an exploration-driven information gain term. Our approach allows the robot to not only actively control its camera for situational awareness for safe navigation, but also to effectively explore the environment. We demonstrated the effectiveness of our framework through extensive studies in simulation and real-world deployments. Our results show that the proposed policy leads to higher success rates, fewer collisions, and improved environment observation compared to static, navigation-only baselines. In future work, we plan to extend this framework to include semantic information and reasoning, enabling the agent to actively search for specific objects of interest. Likewise, we aim to improve generalization to dynamic environments and explore on-the-fly policy adaptation.

\bibliographystyle{IEEEtran}
\bibliography{./bibliography}

\end{document}